\documentclass[lettersize,journal]{IEEEtran}
\usepackage{amsmath,amsfonts}
\usepackage{algorithmic}
\usepackage{algorithm}
\usepackage{array}
\usepackage[caption=false,font=normalsize,labelfont=sf,textfont=sf]{subfig}
\usepackage{textcomp}
\usepackage{stfloats}
\usepackage{url}
\usepackage{verbatim}
\usepackage{graphicx}
\usepackage{cite}
\usepackage[table]{xcolor}
\definecolor{mygreen}{RGB}{0,120,0}
\definecolor{myred}{RGB}{180,0,0}
\definecolor{car}{RGB}{238,59,59}
\definecolor{pedestrian}{RGB}{255,127,0}
\definecolor{rider}{RGB}{255,0,255}
\definecolor{largevehicle}{RGB}{0,255,0}
\usepackage{booktabs}
\usepackage{multirow}
\usepackage{makecell}
\usepackage{threeparttable}
\usepackage{orcidlink}
\usepackage{hyperref}
\hypersetup{hidelinks}

\begin{document}

\title{Sparse4D-Radar: An Efficient and Robust Framework for Surround-View 3D Object Detection via 4D Radar-Camera Fusion}

\author{
	Fuyuan Ai\orcidlink{0009-0005-7944-2913}, Yuchen Tan\orcidlink{0009-0002-6577-5475}, Jiehui Chen\orcidlink{0009-0002-5152-2063}, Zhiwei Xu\orcidlink{0000-0003-2279-0632}, and Chunyi Song\orcidlink{0000-0002-3274-6806} 
\thanks{This work was supported in part by the Fundamental Research Funds for the Central Universities under
	Grant 226202500003. (Corresponding author: Chunyi Song.)}
\thanks{Fuyuan Ai, Yuchen Tan, Jiehui Chen and Zhiwei Xu are are with the State Key Laboratory of Ocean Sensing and Ocean College, Zhejiang University, Zhoushan 316021, China, and also with the Engineering Research Center of Oceanic Sensing Technology and Equipment, Ministry of Education, Zhejiang University, Zhoushan 316021, China (e-mail: \url{fyai@zju.edu.cn}; \url{12334068@zju.edu.cn}; \url{22334009@zju.edu.cn}; \url{xuzw@zju.edu.cn}).}
\thanks{Chunyi Song is with the Donghai Laboratory, Zhoushan 316021, China, also with the Ocean College, Institute of Marine Electronics and Intelligent Systems, Zhejiang University, Zhoushan 316021, China, and also with the Engineering Research Center of Oceanic Sensing Technology and Equipment, Ministry of Education, Zhejiang University, Zhoushan 316021, China (\url{cysong@zju.edu.cn}).}}



\maketitle

\begin{abstract}
In recent years, the rapid advancement of 4D imaging radar has attracted significant attention in autonomous driving due to its superior robustness in adverse weather and its unique capability to provide velocity measurements. However, most existing 4D radar-camera fusion frameworks focus exclusively on front-view perception, leaving a conspicuous gap in surround-view sensing. Moreover, simply extending these methods to 360° coverage often incurs prohibitive computational overhead, hindering their practical deployment. To address these challenges, we propose Sparse4D-Radar, an efficient and robust framework designed for surround-view radar-camera fusion. Specifically, a novel Deformable Fusion module is first developed to seamlessly integrate 4D radar and camera features into the sparse-query mechanism, forming the core of the efficiency-oriented configuration, Sparse4D-Radar-Base. To further push the boundaries of localization precision and modal reliability, two specialized components are incorporated: the Velocity-Consistency Sampling (VCS) module, which leverages velocity measurements for motion-aware feature refinement, and the Adaptive Modality Gating (AMG) module, which dynamically regulates cross-modal integration based on feature reliability. The synergistic integration of these advanced modules culminates in our high-accuracy variant, Sparse4D-Radar-Acc, specifically designed for scenarios requiring maximum detection performance. Extensive evaluations on the OmniHD-Scenes dataset demonstrate that Sparse4D-Radar establishes a new state-of-the-art in surround-view 3D object detection. Our framework significantly outperforms existing methods in diverse environmental conditions, delivering substantial gains in both mAP (over 7\%) and ODS (over 10\%). Notably, it maintains a high inference speed (nearly 10 FPS), achieving an optimal balance between detection accuracy, environmental robustness, and computational efficiency. The code is available at: \url{https://github.com/Aiuan/Sparse4D-Radar}.
\end{abstract}

\begin{IEEEkeywords}
Object detection, 4D radar, camera, fusion perception, real-time, adverse weather.
\end{IEEEkeywords}

\section{Introduction}
\label{sec:int}

\begin{figure}[t]
	\centering
	\includegraphics[width=1.0\linewidth]{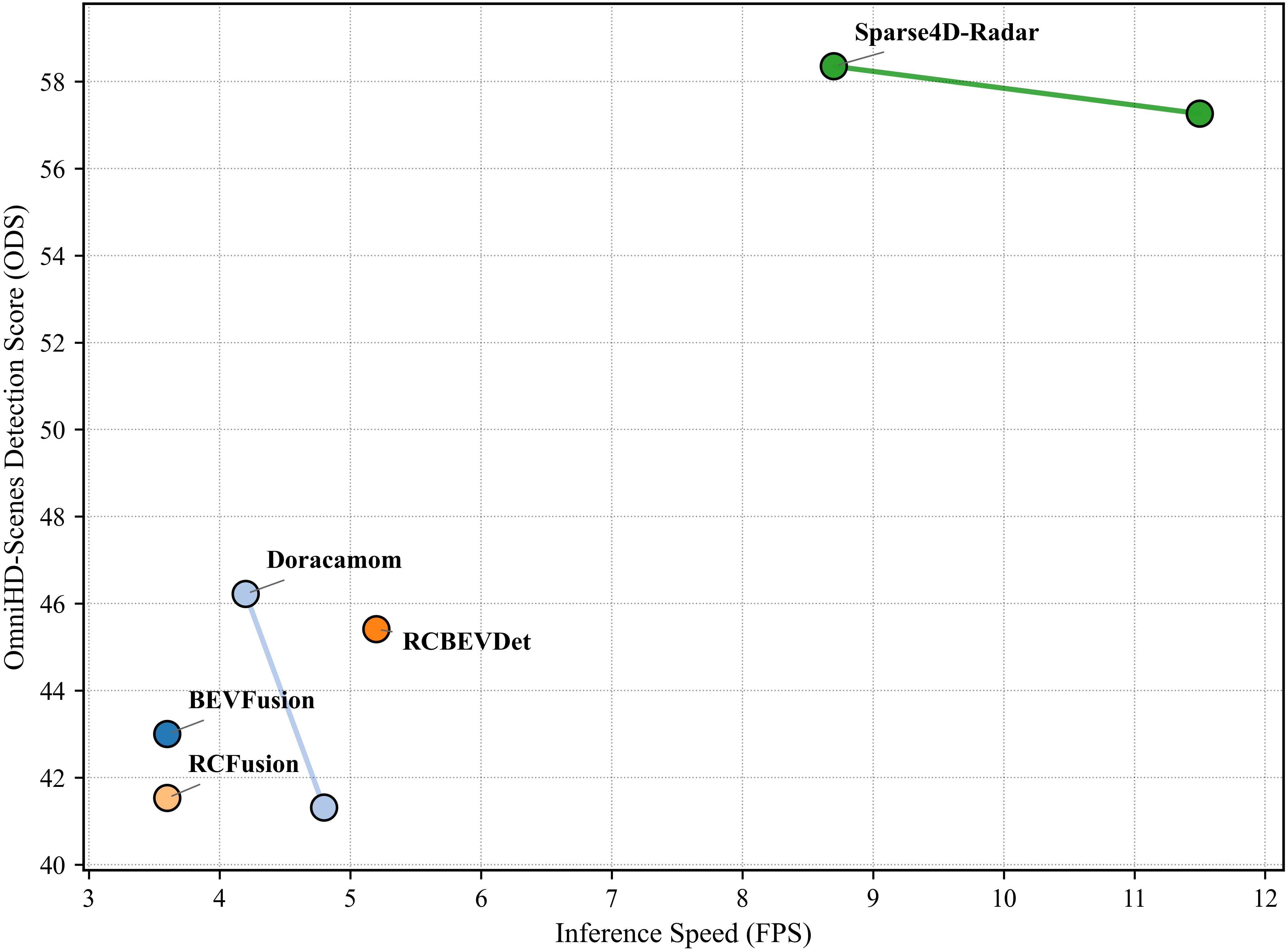}	
	\caption{Comparison of performance and efficiency of various 3D object detection methods on the OmniHD-Scenes test set. The inference speed is measured with a single consumer-grade RTX 4090 GPU. The proposed Sparse4D-Radar achieves the best trade-off between detection accuracy and real-time performance.}
	\label{fig:performance}
\end{figure}

\IEEEPARstart{I}{n} the pursuit of reliable autonomous driving, 4D imaging radar~\cite{fan4DMmWaveRadar2024} has garnered significant traction as a robust candidate for all-weather sensing~\cite{paekKRadar4DRadar2022}. It transcends the limitations of traditional 3D radar by providing high-fidelity spatial information, including crucial elevation data. Compared to vision-based or LiDAR sensors, 4D radar exhibits unparalleled robustness~\cite{chaeRobust3DObject2024} against atmospheric disturbances (e.g., precipitation and obscuration) and offers direct velocity sensing. Despite these merits, the stochastic sparsity of 4D radar data poses challenges for detailed object categorization~\cite{liuSMURFSpatialMultiRepresentation2024a}. This creates a natural synergy with cameras, which provide dense semantic cues but lack depth and weather-resilience~\cite{liuImageAdaptiveYOLOObject2022a}. Consequently, the research community is increasingly focusing on multi-modal fusion frameworks to leverage the strengths of both sensors~\cite{liSemanticguidedDepthCompletion2024,liPCGNetPointCloud2024,songRaViDeepTargetDetection2024,yaoWaterScenesMultiTask4D2024,wangC4RFNetCamera4DRadar2025}.

Within the domain of radar-camera fusion~\cite{yaoRadarCameraFusionObject2024}, 3D object detection serves as a fundamental yet indispensable task for autonomous driving and has witnessed remarkable progress in recent years. However, most existing methods~\cite{zhengRCFusionFusing4D2023,zhangTL4DRCFTwoLevel4D2024a,xiongLXLLiDARExcluded2024b,yangZFusionEffectiveFuser2025,guHGSFusionRadarCameraFusion2025,xiongLXLv2EnhancedLiDAR2025,baiSGDet3DSemanticsGeometry2025,liuMSSF4DRadar2025,zhongCVFusionCrossViewFusion2025,caoDRNetDualRepresentationNetwork2026,xiaR4Det4DRadarCamera2026} remain primarily restricted to front-view perception, lagging behind the prevailing trend toward all-encompassing surround-view understanding—a capability already well-established in vision-only systems. When attempting to extend front-view models to a full 360° field-of-view (FoV), researchers inevitably face a dilemma between perceptual coverage and computational throughput. On one hand, extending the sensing range from a confined front-view to a seamless surround-view leads to an exponential surge in data volume and inference latency. On the other hand, the pervasive reliance on dense Bird’s-Eye-View (BEV) grids for feature alignment imposes a prohibitive memory footprint, particularly when integrating the high-resolution point clouds characteristic of 4D imaging radar. As illustrated in Fig. \ref{fig:performance}, many representative methods are confined to the low-speed region, making them difficult to deploy on resource-constrained onboard platforms. These bottlenecks significantly impede the practical application of high-performance fusion models.

To circumvent these challenges, we propose Sparse4D-Radar, an efficient and robust fusion framework designed for surround-view 3D object detection. Inspired by the success of the Sparse4D~\cite{linSparse4DMultiview3D2023,linSparse4DV2Recurrent2023,linSparse4DV3Advancing2023} series, our framework extends the sparse query-based paradigm to the multi-modal domain, specifically optimized for the integration of 4D radar and camera data. At its core, a novel Deformable Fusion module is developed to bridge the domain gap between heterogeneous modalities; unlike traditional rigid strategies, it employs a deformable sampling mechanism to iteratively align radar spatial cues with camera semantics, effectively resolving sensor misalignment. Building upon this foundation, we further enhance the quality of radar feature by introducing the Velocity-Consistency Sampling (VCS) module. By explicitly leveraging the distinctive velocity dimension of 4D radar, VCS establishes a consistency constraint to compute the velocity similarity between radar points and predicted object. This mechanism effectively identifies and suppresses non-physical noise and ghost reflections that deviate from legitimate object kinematics, ensuring that only motion-compliant radar cues contribute to the subsequent fusion process. To address the fluctuating reliability of sensors across varying environmental conditions, we also incorporate the Adaptive Modality Gating (AMG) module. AMG employs a dynamic gating mechanism to evaluate the real-time confidence of heterogeneous features, such as camera semantics under adverse weather or radar returns in multipath-heavy environments. By adaptively regulating the fusion weights, AMG intelligently balances the information flow between modalities, ensuring that the framework maintains robust and reliable perception even when the performance of a specific sensor is compromised by external noise or occlusions.

Based on these modules, we establish two distinct configurations: Sparse4D-Radar-Base and Sparse4D-Radar-Acc. The Base version serves as an efficiency-oriented baseline, based on the core Deformable Fusion module for high-speed inference. In contrast, the Acc variant incorporates the full suite of components—including VCS and AMG—to push the performance ceiling. Extensive evaluations on the OmniHD-Scenes~\cite{zhengOmniHDScenesNextGenerationMultimodal2026} benchmark demonstrate that the Sparse4D-Radar framework delivers a superior balance of high precision, computational efficiency, and environmental robustness.

The main contributions of this paper are summarized as follows:
\begin{itemize}
	\item \textbf{Novel Framework.} We propose Sparse4D-Radar, an efficient and robust framework for surround-view 3D object detection via radar-camera fusion. It leverages Deformable Fusion as the primary engine for cross-modal interaction, while incorporating the VCS and AMG to enhance radar feature quality and ensure adaptive modality confidence.
	\item \textbf{Scalable Models.} We provide two distinct configurations, Sparse4D-Radar-Base and Sparse4D-Radar-Acc, to accommodate varying computational constraints, where the former serves as an efficiency-oriented baseline for real-time applications and the latter pushes the performance ceiling by incorporating the full suite of refinement modules.
	\item \textbf{State-of-the-Art Performance.} Extensive evaluations on the OmniHD-Scenes benchmark demonstrate the superiority of our approach. Specifically, the Base model achieves 57.25\% ODS (with 47.01\% mAP) while maintaining a high inference speed of 11.5 FPS. The Acc model further improves the performance to 58.35\% ODS (with 47.57\% mAP) at 8.7 FPS. Notably, both configurations exhibit exceptional robustness in adverse scenarios, with the Base and Acc models reaching 58.58\% ODS (with 49.05\% mAP) and 58.61\% ODS (with 48.42\% mAP), respectively, underscoring their reliability for all-weather autonomous perception.
\end{itemize}

\section{Related Work}
\label{sec:rel}

\subsection{Radar-Camera Fusion for 3D Object Detection}

The synergy between 4D radar and camera has emerged as a pivotal research direction for robust 3D perception. Early fusion strategies, such as RCFusion~\cite{zhengRCFusionFusing4D2023} and adapted versions of BEVFusion~\cite{liuBEVFusionMultiTaskMultiSensor2023a}, primarily focus on the direct concatenation of radar and camera features within the BEV space. These aggregated features are subsequently processed by a convolution-based encoder to generate fusion features. Although computationally efficient and straightforward, these methods often suffer from suboptimal performance due to the inherent spatial misalignment between heterogeneous features originating from separate backbones.

To resolve such misalignment issues, subsequent research began leveraging radar as a geometric guide to mitigate the depth ambiguity inherent in lifting perspective-view (PV) features into BEV space. For instance, LXL~\cite{xiongLXLLiDARExcluded2024b,xiongLXLv2EnhancedLiDAR2025} utilizes 3D radar occupancy grids to assist depth sampling, while CRN~\cite{kimCRNCameraRadar2023b} and SGDet3D~\cite{baiSGDet3DSemanticsGeometry2025} incorporate sparse radar depth cues to supervise image-based depth prediction, thereby enhancing the spatial fidelity of the resulting representations. Besides these geometric constraints, more sophisticated attention mechanisms have been introduced to foster deeper cross-modal interaction and further address spatial dislocation. CRN~\cite{kimCRNCameraRadar2023b} and RCBEVDet~\cite{linRCBEVDetRadarCameraFusion2024a,linRCBEVDetHighaccuracyRadarCamera2024} employ deformable cross-attention to dynamically align features, allowing the model to adaptively focus on relevant regions across different modalities.

In parallel with these fusion-level improvements, recent studies have also highlighted that the quality of raw radar input is equally critical. For example, RCBEVDet~\cite{linRCBEVDetRadarCameraFusion2024a,linRCBEVDetHighaccuracyRadarCamera2024} proposes a dual-branch backbone to refine radar features, while HGSFusion~\cite{guHGSFusionRadarCameraFusion2025} utilizes instance segmentation masks to generate high-fidelity hybrid point clouds. Meanwhile, the effective exploitation of radar velocity attributes has also emerged as an important direction. DR-Net~\cite{caoDRNetDualRepresentationNetwork2026} introduces a dual-representation encoder and motion-aware augmentation to better leverage velocity cues and mitigate feature information loss in 4D radar-based detection. 

Despite these advancements, most BEV-based paradigms still grapple with the heavy computational burden of dense grids and error-prone implicit depth estimation. Consequently, the field is witnessing a strategic shift toward sparse query-based architectures. Frameworks such as RaCFormer~\cite{chuRaCFormerHighQuality3D2025} and SpaRC~\cite{woltersSpaRCSparseRadarCamera2025a} leverage the inherent efficiency of sparse queries to achieve superior alignment without the need for dense transformations. Aligning with this cutting-edge trajectory, our proposed Sparse4D-Radar further optimizes the sparse fusion process, specifically tailored to exploit the unique multi-dimensional characteristics of 4D imaging radar.

\subsection{Surround-view 3D Object Detection}

The evolution of surround-view 3D object detection has been closely intertwined with the advancement of multi-modal datasets and sensor technologies. Previously, numerous frameworks~\cite{kimCRNCameraRadar2023b,linRCBEVDetRadarCameraFusion2024a,linRCBEVDetHighaccuracyRadarCamera2024,chuRaCFormerHighQuality3D2025,woltersSpaRCSparseRadarCamera2025a} leveraging traditional 3D radar and camera fusion have achieved commendable surround-view performance, with the majority of these efforts validated on the nuScenes~\cite{caesarNuScenesMultimodalDataset2020} benchmark. However, this perception paradigm is inherently constrained by the physical limitations of 3D radar, such as the lack of elevation information and low point cloud density, which creates a significant performance bottleneck for high-precision localization. 

The emergence of 4D imaging radar offered a promising solution to these bottlenecks; however, it rendered many conventional 3D radar-based methodologies obsolete. To facilitate algorithmic research in this new domain, several 4D radar datasets, such as VoD~\cite{palffyMultiClassRoadUser2022a}, TJ4DRadSet~\cite{zhengTJ4DRadSet4DRadar2022a}, and K-Radar~\cite{paekKRadar4DRadar2022}, are subsequently released. However, these early benchmarks primarily provided front-view sensor data, thereby severely restricting the exploration of all-around perception.

The introduction of the OmniHD-Scenes~\cite{zhengOmniHDScenesNextGenerationMultimodal2026} dataset has recently alleviated this data scarcity by offering a comprehensive multi-sensor configuration for surround-view perception. Nevertheless, dedicated algorithms designed for this benchmark remain scarce, with most existing baselines being merely migrated from prior front-view models, leading to suboptimal accuracy in complex 360° scenarios. To address this gap, Doracamom~\cite{zhengDoracamomJoint3D2026} introduces a multi-task joint perception framework that utilizes both voxel and BEV representations for comprehensive spatio-temporal learning, effectively fusing multi-modal features to achieve high-quality 3D object detection and semantic occupancy prediction. Moreover, RaGS~\cite{baiRaGSUnleashing3D2025} leverages 3D Gaussian Splatting to model the scene as a continuous field, employing a cascaded pipeline to progressively refine object-centric features by integrating camera semantics with 4D radar velocity and geometry.

Aligning with this trajectory of FoV-centric evolution, we propose Sparse4D-Radar, a dedicated surround-view framework tailored for 4D radar-camera fusion. Our approach focuses on optimizing the sparse interaction between multi-view visual cues and the multi-dimensional radar attributes, ensuring robust and precise 3D object detection across the entire 360° coverage.

\subsection{Efficient and Real-time 3D Object Detection}

Considering the practical deployment in autonomous driving, the computational efficiency of perception algorithms remains as critical as their detection accuracy. This necessity is primarily driven by the stringent computational constraints of onboard processing units, which must handle massive multi-sensor data streams in real-time with limited power consumption and memory bandwidth. In this context, earlier fusion methodologies~\cite{zhengRCFusionFusing4D2023,liuBEVFusionMultiTaskMultiSensor2023a,linRCBEVDetRadarCameraFusion2024a,zhengDoracamomJoint3D2026} relying on dense BEV representations have encountered significant scalability challenges. In these paradigms, the number of queries and keys typically scales with the total number of BEV grid cells, leading to an explosive growth in computational complexity during cross-modal feature aggregation. Even with the integration of deformable attention~\cite{zhuDeformableDETRDeformable2021} to sample a limited number of keys, the underlying overhead of generating and maintaining dense spatial features remains a substantial barrier to achieving high-frequency real-time inference.

To mitigate these efficiency bottlenecks, sparse-query based architectures~\cite{chuRaCFormerHighQuality3D2025,woltersSpaRCSparseRadarCamera2025a} have increasingly garnered research attention as a compelling alternative. By representing the scene with a predefined set of learned object queries rather than exhaustive spatial grids, these methods significantly reduce the redundant computation inherent in dense sampling while maintaining commendable detection performance. This shift toward sparsity-driven interaction has established a new frontier for balancing latency and precision. Inheriting the architectural elegance of such sparse paradigms, our proposed Sparse4D-Radar preserves these inherent advantages in computational agility, while specifically focusing on strengthening the cross-modal synergy to further elevate the accuracy and robustness of radar-camera fusion.

\section{Methodology}
\label{sec:met}

\subsection{Overall Framework}

\begin{figure*}[t]
	\centering
	\includegraphics[width=1.0\linewidth]{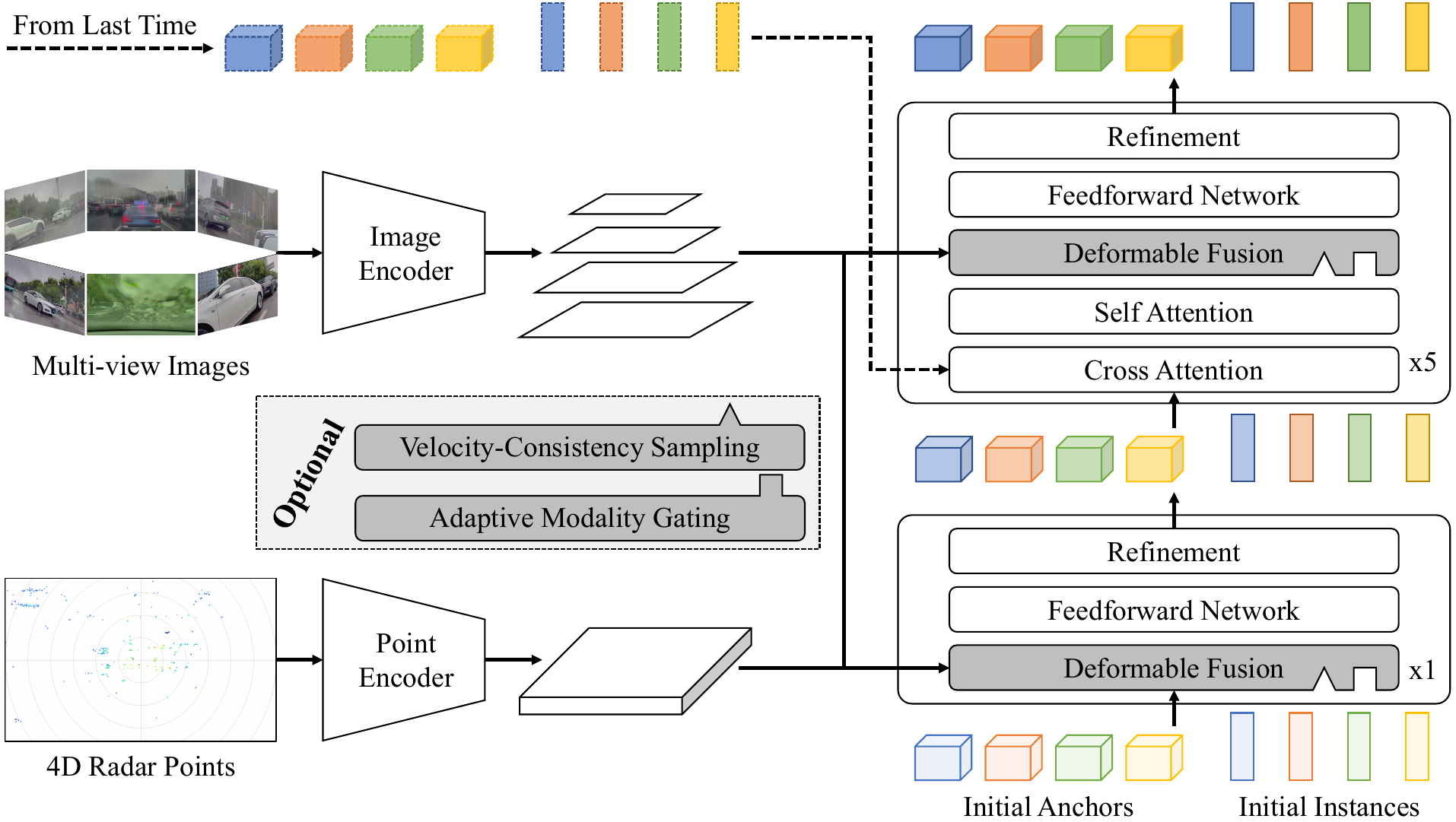}	
	\caption{The overall framework of \textbf{Sparse4D-Radar}. Multi-view images and 4D radar points are processed by dedicated encoders to extract multi-modal features. The anchors and instances are refined through cascaded decoder blocks to generate final detection results. At the core of the framework, the Deformable Fusion module enables effective multi-modal feature aggregation. Built upon this foundation, the VCS and AMG modules are introduced as plug-and-play components to further enhance sampling accuracy and robustness under challenging environmental conditions, respectively.}
	\label{fig:framework}
\end{figure*}

As illustrated in Fig. 2, the proposed Sparse4D-Radar is an end-to-end framework for surround-view 3D object detection, which extends the sparse query-based paradigm to a multi-modal setting by integrating 4D radar and camera data.

The Image Encoder is employed to extract multi-view image features using shared weights, comprising both a backbone (e.g., ResNet~\cite{heDeepResidualLearning2016}) and a neck (e.g., FPN~\cite{linFeaturePyramidNetworks2017}). Given $N$ camera views of the current frame, it generates multi-view and multi-scale PV feature maps. In parallel, the Point Encoder processes 4D radar point clouds. By utilizing a backbone (e.g., PointPillars~\cite{langPointPillarsFastEncoders2019a}) and a neck (e.g., SECONDFPN~\cite{yanSECONDSparselyEmbedded2018a}), it transforms raw points into multi-scale BEV feature maps.

The extracted multi-modal features, together with a set of initialized 3D anchors and instance features, are then fed into a series of cascaded decoder blocks for iterative refinement. While the initial decoder block is simplified for efficient priming, the subsequent blocks include consistent and comprehensive modules to refine the detection results. 

Within each block, the Cross-Attention module aggregates temporal information from historical instances, enabling temporal interaction across frames. The Self-Attention module models inter-instance relationships to capture spatial dependencies and suppress redundant predictions. The core feature aggregation is performed by the Deformable Fusion module, which adaptively integrates radar geometric cues and visual semantic features conditioned on the evolving queries. To further enhance localization accuracy and modality robustness, two optional components can be incorporated into this module: the VCS module exploits radar velocity measurements to enforce motion-aware feature aggregation, and the AMG module dynamically adjusts the contribution of each modality based on its reliability under varying environmental conditions. These components can be flexibly enabled to strengthen the model’s performance in challenging scenarios. Following feature fusion, a Feedforward Network is applied to enhance the representation capacity of instance features, and a Refinement module predicts classification scores and regresses 3D bounding box attributes.

To maintain temporal consistency, the outputs of the final decoder block are cached and propagated to subsequent frames as temporal priors. This recurrent design allows the model to leverage long-term temporal cues and motion dynamics.

\subsection{Deformable Fusion}

\begin{figure*}[t]
	\centering
	\includegraphics[width=1.0\linewidth]{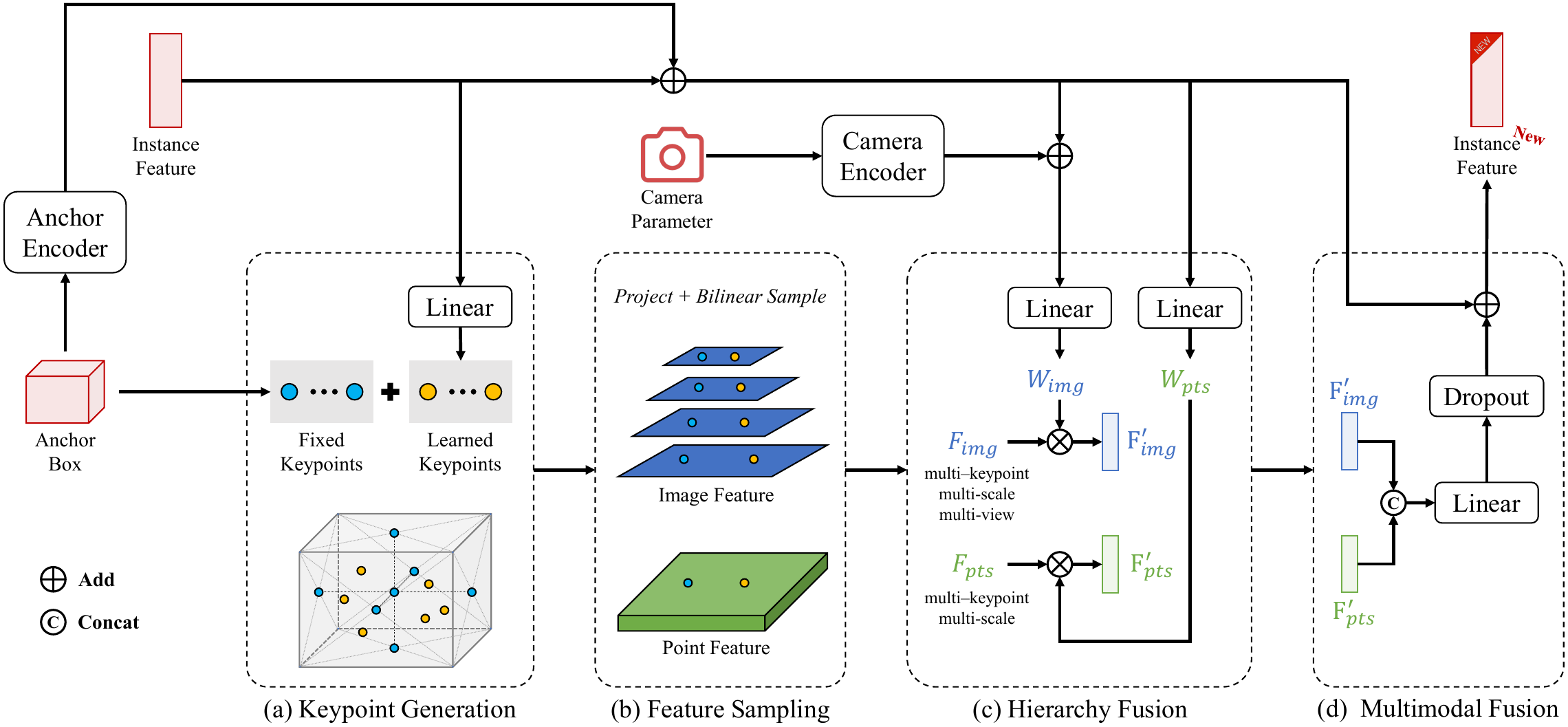}	
	\caption{Illustration of the proposed \textbf{Deformable Fusion} module. High-quality feature extraction is achieved in four stages: (a) Keypoint Generation, producing fixed and learned 3D keypoints from each anchor box; (b) Feature Sampling, , projecting keypoints and sampling features from multi-view/scale image and point maps; (c) Hierarchy Fusion, adaptively aggregating features across scales, views, and keypoints for each object; and (d) Multimodal Fusion, integrating refined features from disparate modalities to update the instance representation.}
	\label{fig:deformablefusion}
\end{figure*}

As the architectural cornerstone, the Deformable Fusion module enables the extraction and integration of multi-modal features for a specific set of anchors and instances. As illustrated in Fig.~\ref{fig:deformablefusion}, this mechanism includes four critical steps:

\textbf{Keypoint Generation.} Following Sparse4D~\cite{linSparse4DMultiview3D2023, linSparse4DV2Recurrent2023, linSparse4DV3Advancing2023}, we generate a set of representative 3D keypoints for each query to capture its spatial cues for subsequent feature extraction. A set of fixed keypoints is first established based on the locations and dimensions of the anchors, providing a fundamental geometric prior. Recognizing that the optimal sampling regions vary across different object categories and shapes, learnable keypoints are further incorporated to provide spatial flexibility. These points are predicted from the corresponding instance features via a simple linear layer, allowing the model to adaptively shift sampling locations toward the most informative parts of a specific object.

\textbf{Feature Sampling.} Once the keypoints are generated, they are projected into the respective coordinate spaces of the images and the point clouds. For a keypoint $\mathbf{P}_{key}$, its projected location $\mathbf{P}_{img}$ in the image PV space and its corresponding position $\mathbf{P}_{pts}$ in the radar BEV space are determined as follows:
\begin{align}	
	\mathbf{P}_{img} & = \mathbf{K}_n^{cam}[\mathbf{R}_n^{cam}|\mathbf{t}_n^{cam}]\mathbf{P}_{key},
	\label{eq1} \\
	\mathbf{P}_{pts} & = [\mathbf{R}^{radar}|\mathbf{t}^{radar}]\mathbf{P}_{key},
	\label{eq2}
\end{align}
where $n = 1, \dots, N$ denotes the $n$-th camera index, while $\mathbf{K}$, $\mathbf{R}$, and $\mathbf{t}$ represent intrinsic matrix, rotation matrix, and translation vector, respectively. Based on these projected coordinates, we employ bilinear interpolation to extract features from the image feature maps and point feature maps, resulting in the keypoint-wise features $\mathbf{F}_{img}$ and $\mathbf{F}_{pts}$.

\textbf{Hierarchy Fusion.} To aggregate the multi-level features—where $\mathbf{F}_{img}$ contains information across multiple keypoints, scales, and views, and $\mathbf{F}_{pts}$ covers multiple keypoints and scales—we employ a learnable weighting mechanism to fuse these features on a per-modality basis. The learnable weights $\mathbf{W}_{img}$ and $\mathbf{W}_{pts}$ are computed as:
\begin{align}	
	\mathbf{W}_{img} & = MLP(\mathbf{F}_{ins}+\mathbf{E}_{anc}+\mathbf{E}_{cam}),
	\label{eq3} \\
	\mathbf{W}_{pts} & = MLP(\mathbf{F}_{ins}+\mathbf{E}_{anc}),
	\label{eq4}
\end{align}
where $\mathbf{F}_{ins}$ and $\mathbf{E}_{anc}$ represent the instance feature and anchor embedding, respectively, and $\mathbf{E}_{cam}$ is the camera embedding encoding the specific sensor parameters. Subsequently, an element-wise product is performed between $\mathbf{F}_{img}$ and $\mathbf{W}_{img}$, followed by a summation across the keypoint, scale, and view dimensions to obtain the compressed feature $\mathbf{F}_{img}'$. Likewise, $\mathbf{F}_{pts}$ is multiplied by $\mathbf{W}_{pts}$ and aggregated across keypoints and scales to yield $\mathbf{F}_{pts}'$.

\textbf{Multimodal Fusion.} The obtained $\mathbf{F}_{img}'$ and $\mathbf{F}_{pts}'$ are concatenated and passed through a linear layer and a dropout layer. The resulting output is then utilized to update the instance feature, facilitating the integration of complementary information from both camera and radar modalities.

\subsection{Velocity-Consistency Sampling}

\begin{figure}[t]
	\centering
	\includegraphics[width=1.0\linewidth]{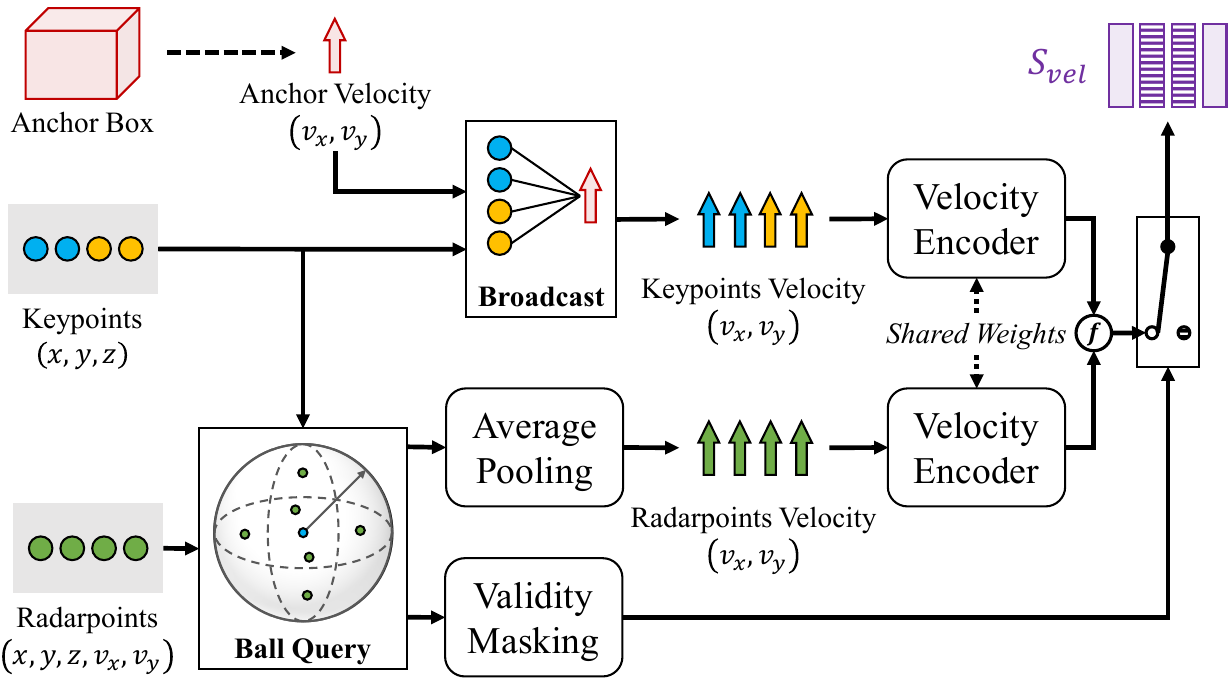}	
	\caption{Illustration of the proposed \textbf{Velocity-Consistency Sampling} module.}
	\label{fig:velocityconsistencysampling}
\end{figure}

To address the challenges posed by noise and ghost points in radar point clouds, which often hinder sampling accuracy, we introduce the VCS module, as illustrated in Fig. \ref{fig:velocityconsistencysampling}. Initially, the velocity of the anchor box is extracted and broadcasted to the corresponding keypoints $\mathbf{P}_{key}$, resulting in the keypoint velocity. Simultaneously, neighboring radar points $\mathbf{P}_{rad}$ for each keypoint are retrieved via ball query~\cite{qiPointNetDeepLearning2017,qiPointNetDeepHierarchical2017a}. The velocities of these neighboring radar points are then processed through average pooling operation to compute the neighborhood radar velocity. Both the keypoint and neighborhood radar velocities are passed through a shared-weight velocity encoder to generate velocity embeddings $\mathbf{E}_{vel}^{key}$ and $\mathbf{E}_{vel}^{rad}$, which are then used to compute the velocity similarity $S_{vel}$ as follow:
\begin{align}
	\mathbf{S}_{vel} & = 
	\begin{cases}
		\sigma(\sum \frac{\mathbf{E}_{vel}^{key} \odot \mathbf{E}_{vel}^{rad}}{\sqrt{C}}), & \|\mathbf{P}_{key} - \mathbf{P}_{rad}\|_2 \leq r\\
		s, & otherwise,
	\end{cases}
	\label{eq5}
\end{align}
where $\sigma$ is the sigmoid activation function, $\odot$ denotes the element-wise multiplication, $C$ represents the number of channels in the velocity encoder, $\mathbb{I}$ is the indicator function that is 1 when the condition inside the parentheses holds true and 0 otherwise, $\|\:\|_2$ represents the Euclidean distance between points, $r$ is the radius of the neighborhood used in the ball query, and $s$ denotes the default similarity for keypoints that have no neighboring radar points. In this work, both $r$ and $s$ are empirically set to 0.5 to balance local geometric capture and feature stability.

The resulting similarity score $S_{vel}$ can be seamlessly integrated into the Deformable Fusion process by performing an element-wise product with $\mathbf{F}_{pts}$ along the keypoint dimension, effectively recalibrating the radar point features to favor object with consistent motion dynamics.

\subsection{Adaptive Modality Gating}

\begin{figure}[t]
	\centering
	\includegraphics[width=1.0\linewidth]{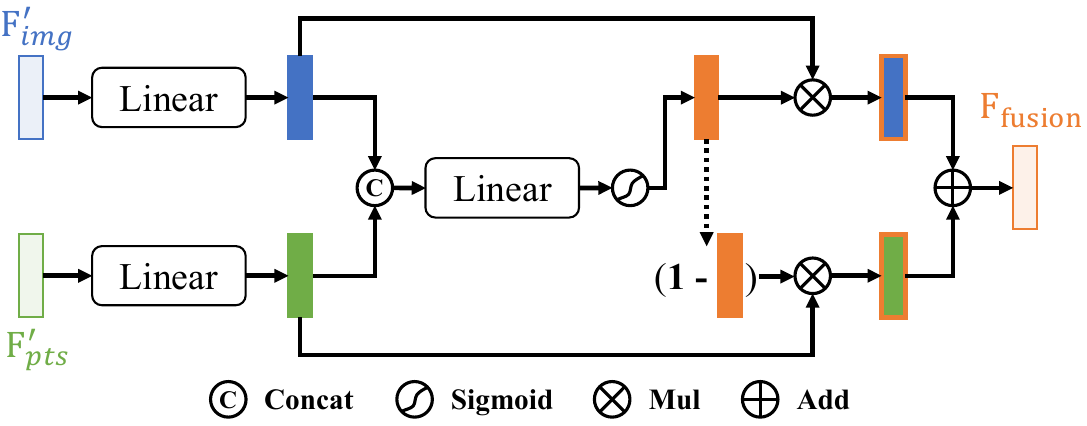}	
	\caption{Illustration of the proposed \textbf{Adaptive Modality Gating} module.}
	\label{fig:adaptivemodalitygating}
\end{figure}

To account for the varying confidence levels of different modalities across diverse environments (e.g., image degradation in rainy or foggy weather), we introduce the AMG module, as illustrated in Fig. \ref{fig:adaptivemodalitygating}. This module employs a gating mechanism to adaptively adjust the contribution of each modality based on its perceived reliability, formulated as:
\begin{align}	
	\mathbf{G} & = \sigma(MLP([\mathbf{F}_{img}'';\mathbf{F}_{pts}''])),
	\label{eq6} \\
	\mathbf{F}_{fusion} & = MLP(\mathbf{G} \odot \mathbf{F}_{img}'' + (1 - \mathbf{G}) \odot \mathbf{F}_{pts}''),
	\label{eq7}
\end{align}
where $\mathbf{F}_{img}''$ and $\mathbf{F}_{pts}''$ are the linear projections of $\mathbf{F}_{img}'$ and $\mathbf{F}_{pts}'$, respectively. Here, $\sigma$ represents the sigmoid activation function, while $\odot$ denotes the element-wise multiplication.

This mechanism can be easily used in the multi-modal fusion step of Deformable Fusion. By replacing the original concatenated features with the obtained fusion features $\mathbf{F}_{fusion}$, a more robust instance feature can be generated. When the image quality is degraded (e.g., due to noise or occlusion), the gating network automatically decreases $\mathbf{G}$, thereby suppressing the unreliable image features and relying more on the robust geometric information from the point clouds. This adaptive behavior ensures the model's robustness across diverse and challenging scenarios.

\subsection{Detection Head and Loss Function}

Following DETR~\cite{carionEndtoEndObjectDetection2020b,wangDETR3D3DObject2022a}, our detection head is structured as a transformer-based decoder that iteratively refines a set of sparse, learnable object queries. To enable end-to-end 3D object detection without post-processing, we employ a set-to-set loss function utilizing Hungarian Matching to establish a unique one-to-one assignment between predictions and ground truth. The total objective is decomposed into detection-specific supervision and auxiliary depth constraints to facilitate the geometric perception of the image branch.

The detection loss $\mathcal{L}_{det}$ consists of a classification term and a regression term. For classification ($\mathcal{L}_{cls}$), Focal Loss~\cite{linFocalLossDense2017} is adopted to alleviate the class imbalance. For regression ($\mathcal{L}_{reg}$), L1 loss is employed to supervise the 3D bounding box attributes, including coordinates, dimensions, orientation (yaw), and velocity. For depth supervision, the depth loss $\mathcal{L}_{depth}$ is defined as Masked L1 loss, which computes the absolute error between the predicted depth map and the sparse depth map obtained via LiDAR projection. Finally, the total loss $\mathcal{L}_{total}$ is formulated as a weighted summation of these components:
\begin{align}	
	\mathcal{L}_{total} & = \lambda_{cls}\mathcal{L}_{cls} + \lambda_{reg}\mathcal{L}_{reg} + \lambda_{dep}\mathcal{L}_{depth},
	\label{eq8}
\end{align}
where $\lambda_{cls}$, $\lambda_{reg}$, and $\lambda_{dep}$ are the loss weights used to balance the contribution of each task. In this work, the default values for these hyperparameters are set to 2.0, 0.25, and 0.2, respectively.

\section{Experiments}
\label{sec:exp}

\subsection{Dataset and Evaluation Metrics}

\textbf{Dataset.} We evaluate our method on the OmniHD-Scenes~\cite{zhengOmniHDScenesNextGenerationMultimodal2026} dataset, the first large-scale multi-view perception benchmark specifically designed for 4D radar-based sensing. The sensor suite comprises six surround-view cameras, six surround-view 4D millimeter-wave radars, and a high-resolution 128-beam LiDAR. The dataset consists of 1,501 clips captured across diverse weather conditions and time periods. For the 3D object detection task, 200 clips are meticulously annotated, covering four categories: car, pedestrian, rider, and large\_vehicle. The annotated data includes 11,921 keyframes, which are split into 8,321 frames for training and 3,600 frames for testing.

\textbf{Evaluation Metrics.} To ensure a fair comparison with established baselines, we strictly follow the official evaluation protocol, including both the predefined detection range (±60 m longitudinally and ±40 m laterally relative to the ego vehicle) and the standardized metrics. Under this unified setting, we report the widely adopted mean Average Precision (mAP) as a base metric, together with four mean True Positive (mTP) metrics—mean Average Translation Error (mATE), mean Average Scale Error (mASE), mean Average Orientation Error (mAOE), and mean Average Velocity Error (mAVE)—to provide a more fine-grained analysis of detection quality. More importantly, we adopt the OmniHD-Scenes Detection Score (ODS) as the primary evaluation metric, as it offers a more comprehensive assessment of overall detection performance by jointly considering multiple aspects of prediction quality.

\subsection{Implementation Details}

All images are standardized to a resolution of $544 \times 960$ pixels through unified cropping and scaling. To enhance the density of the 4D radar data, we accumulate points across three consecutive frames. Each radar point is represented as a multidimensional feature vector containing $[x, y, z, v_x, v_y, power, snr, \Delta t]$. The spatial extent of the point cloud is constrained to $[-60, 60]$m, $[-40, 40]$m, and $[-3, 5]$m along the X-, Y-, and Z-axes, respectively.

The entire framework is implemented using PyTorch and trained on NVIDIA RTX A6000 GPUs. With the exception of the image encoder, which utilizes pre-trained weights, all modules are initialized from scratch and trained for 100 epochs. We employ the AdamW optimizer with an initial learning rate of $6 \times 10^{-5}$, following a Cosine Annealing scheduler for dynamic learning rate adjustment. This configuration ensures stable convergence while maximizing the synergistic representation learning between the radar and camera modalities.

\subsection{Comparison with Existing Methods}

\begin{table*}[t]
	\centering
	\caption{Comparison of 3D object detection on the OmniHD-Scenes test set.}
	\label{tab:all_conditions}	
	\begin{tabular}{c|c|c|c|cc|cccc}
	\toprule
	\textbf{Method} & \textbf{Modality} & \textbf{Image Res.} & \textbf{Backbone} & \textbf{mAP$\uparrow$} & \textbf{ODS$\uparrow$} 
	& \textbf{mATE$\downarrow$} & \textbf{mASE$\downarrow$} & \textbf{mAOE$\downarrow$} & \textbf{mAVE$\downarrow$} \\
	\midrule
	PointPillars~\cite{langPointPillarsFastEncoders2019a} & L-128 & - & PointPillars & 61.15 & 55.54 & 0.2825 & 0.1980 & 0.5223 & 1.8763 \\
	PointPillars~\cite{langPointPillarsFastEncoders2019a} & L-32 & - & PointPillars & 57.24 & 52.66 & 0.3040 & 0.2038 & 0.5692 & 1.8731 \\
	\midrule
	PointPillars~\cite{langPointPillarsFastEncoders2019a} & R & - & PointPillars & 23.82 & 37.21 & 0.6752 & 0.2447 & 0.3776 & 0.6789 \\
	RadarPillarNet~\cite{zhengRCFusionFusing4D2023} & R & - & RadarPillarNet & 24.88 & 37.81 & 0.6597 & 0.2389 & 0.3736 & 0.6982 \\
	\midrule
	LSS~\cite{philionLiftSplatShoot2020b} & C & 544$\times$960 & R50 & 22.44 & 26.01 & 1.0238 & 0.2230 & 0.5942 & 2.0138 \\
	BEVFormer-S~\cite{liBEVFormerLearningBirdsEyeView2022} & C & 544$\times$960 & R50 & 26.49 & 28.10 & 1.1430 & 0.2315 & 0.5799 & 1.6666 \\
	BEVFormer~\cite{liBEVFormerLearningBirdsEyeView2022} & C & 544$\times$960 & R50 & 29.17 & 30.54 & 1.1046 & 0.2346 & 0.4889 & 1.0797 \\
	Sparse4Dv3~\cite{linSparse4DV3Advancing2023} & C & 544$\times$960 & R50 & 28.65 & 32.99 & 0.9538 & 0.2139 & 0.5575 & 0.7817 \\
	BEVFormer-S~\cite{liBEVFormerLearningBirdsEyeView2022} & C & 864$\times$1536 & R101-DCN & 30.10 & 30.55 & 1.0633 & 0.2266 & 0.5331 & 1.6625 \\
	BEVFormer~\cite{liBEVFormerLearningBirdsEyeView2022} & C & 864$\times$1536 & R101-DCN & 32.22 & 32.57 & 1.0637 & 0.2271 & 0.4558 & 1.0683 \\
	\midrule
	BEVFusion~\cite{liuBEVFusionMultiTaskMultiSensor2023a} & R+C & 544$\times$960 & R50+PointPillars & 33.95 & 43.00 & 0.5703 & 0.2165 & 0.3814 & 0.7474 \\
	RCFusion~\cite{zhengRCFusionFusing4D2023} & R+C & 544$\times$960 & R50+RadarPillarNet & 34.88 & 41.53 & 0.5676 & 0.2135 & 0.3711 & 0.9208 \\
	RCBEVDet~\cite{linRCBEVDetRadarCameraFusion2024a} & R+C & 544$\times$960 & R50+RadarBEVNet & 35.53 & 45.41 & 0.5138 & 0.2305 & 0.3614 & 0.6825 \\
	RaGS~\cite{baiRaGSUnleashing3D2025} & R+C & 544$\times$960 & R50+RadarPillarNet & 35.88 & 43.45 & - & - & - & - \\
	Doracamom-S~\cite{zhengDoracamomJoint3D2026} & R+C & 544$\times$960 & R50+RadarPillarNet & 37.60 & 41.31 & 0.6724 & 0.2329 & 0.4359 & 0.8579 \\
	Doracamom~\cite{zhengDoracamomJoint3D2026} & R+C & 544$\times$960 & R50+RadarPillarNet & 39.12 & 46.22 & 0.6646 & 0.2331 & 0.3545 & 0.6151 \\
	\rowcolor{gray!20} 
	Sparse4D-Radar-Base(\textbf{ours}) & R+C & 544$\times$960 & R50+PointPillars & \underline{47.01} & \underline{57.25} & \underline{0.4388} & \underline{0.1981} & \underline{0.3266} & \underline{0.3368} \\
	\rowcolor{gray!20} 
	Sparse4D-Radar-Acc(\textbf{ours}) & R+C & 544$\times$960 & R50+PointPillars & \textbf{47.57} & \textbf{58.35} & \textbf{0.4199} & \textbf{0.1927} & \textbf{0.3034} & \textbf{0.3187} \\
	\bottomrule
	\multicolumn{10}{l}{\scriptsize \textit{Note:} L-128/32 denote 128-line LiDAR and its manually downsampled 32-line version; R and C represent 4D radar and camera.}
	\end{tabular}	
\end{table*}

\begin{table*}[t]
	\centering
	\caption{Comparison of 3D object detection under bad conditions (night and rainy weather) on the OmniHD-Scenes test set.}
	\label{tab:bad_conditions}	
	\begin{tabular}{c|c|c|c|cc|cccc}
		\toprule
		\textbf{Method} & \textbf{Modality} & \textbf{Image Res.} & \textbf{Backbone} & \textbf{mAP$\uparrow$} & \textbf{ODS$\uparrow$} 
		& \textbf{mATE$\downarrow$} & \textbf{mASE$\downarrow$} & \textbf{mAOE$\downarrow$} & \textbf{mAVE$\downarrow$} \\
		\midrule
		PointPillars~\cite{langPointPillarsFastEncoders2019a} & L-128 & - & PointPillars & 60.02 & 55.03 & 0.2889 & 0.2008 & 0.5084 & 1.8996 \\
		PointPillars~\cite{langPointPillarsFastEncoders2019a} & L-32 & - & PointPillars & 57.09 & 52.94 & 0.3014 & 0.2098 & 0.5370 & 1.9397 \\
		\midrule
		PointPillars~\cite{langPointPillarsFastEncoders2019a} & R & - & PointPillars & 28.63 & 40.70 & 0.6539 & 0.2538 & 0.3703 & 0.6112 \\
		RadarPillarNet~\cite{zhengRCFusionFusing4D2023} & R & - & RadarPillarNet & 29.78 & 41.00 & 0.6473 & 0.2482 & 0.3707 & 0.6447 \\
		\midrule
		LSS~\cite{philionLiftSplatShoot2020b} & C & 544$\times$960 & R50 & 21.21 & 25.62 & 1.0723 & 0.2373 & 0.5615 & 1.8993 \\
		BEVFormer-S~\cite{liBEVFormerLearningBirdsEyeView2022} & C & 544$\times$960 & R50 & 24.57 & 27.25 & 1.1603 & 0.2449 & 0.5581 & 1.6911 \\
		BEVFormer~\cite{liBEVFormerLearningBirdsEyeView2022} & C & 544$\times$960 & R50 & 27.50 & 29.92 & 1.1038 & 0.2477 & 0.4588 & 1.0685 \\
		Sparse4Dv3~\cite{linSparse4DV3Advancing2023} & C & 544$\times$960 & R50 & 26.54 & 32.06 & 0.9952 & 0.2280 & 0.4909 & 0.7829 \\
		BEVFormer-S~\cite{liBEVFormerLearningBirdsEyeView2022} & C & 864$\times$1536 & R101-DCN & 28.11 & 29.36 & 1.0972 & 0.2400 & 0.5358 & 1.6947 \\
		BEVFormer~\cite{liBEVFormerLearningBirdsEyeView2022} & C & 864$\times$1536 & R101-DCN & 30.39 & 31.61 & 1.0613 & 0.2392 & 0.4475 & 1.0780 \\
		\midrule
		BEVFusion~\cite{liuBEVFusionMultiTaskMultiSensor2023a} & R+C & 544$\times$960 & R50+PointPillars & 35.83 & 44.95 & 0.5665 & 0.2317 & 0.3739 & 0.6653 \\
		RCFusion~\cite{zhengRCFusionFusing4D2023} & R+C & 544$\times$960 & R50+RadarPillarNet & 36.36 & 43.29 & 0.5508 & 0.2283 & 0.3616 & 0.8504 \\
		RCBEVDet~\cite{linRCBEVDetRadarCameraFusion2024a} & R+C & 544$\times$960 & R50+RadarBEVNet & 37.49 & 47.32 & - & - & - & - \\
		Doracamom-S~\cite{zhengDoracamomJoint3D2026} & R+C & 544$\times$960 & R50+RadarPillarNet & 38.75 & 43.47 & - & - & - & - \\
		Doracamom~\cite{zhengDoracamomJoint3D2026} & R+C & 544$\times$960 & R50+RadarPillarNet & 41.86 & 48.74 & 0.6315 & 0.2391 & 0.3449 & 0.5596 \\
		\rowcolor{gray!20} 
		Sparse4D-Radar-Base(\textbf{ours}) & R+C & 544$\times$960 & R50+PointPillars & \textbf{49.05} & \underline{58.58} & \underline{0.4524} & \underline{0.2065} & \underline{0.2987} & \underline{0.3175} \\
		\rowcolor{gray!20} 
		Sparse4D-Radar-Acc(\textbf{ours}) & R+C & 544$\times$960 & R50+PointPillars & \underline{48.42} & \textbf{58.61} & \textbf{0.4373} & \textbf{0.2021} & \textbf{0.2979} & \textbf{0.3104} \\		
		\bottomrule
		\multicolumn{10}{l}{\scriptsize \textit{Note:} L-128/32 denote 128-line LiDAR and its manually downsampled 32-line version; R and C represent 4D radar and camera.}
	\end{tabular}	
\end{table*}

\begin{table*}[t]
	\centering
	\caption{Comparison of inference speed and computational cost.}
	\label{tab:computational_cost}	
	\begin{tabular}{c|c|c|c|ccc}
		\toprule
		\textbf{Method} & \textbf{Modality} & \textbf{Image Res.} & \textbf{Backbone} & \textbf{FPS$\uparrow$} & \textbf{FLOPs$\downarrow$} & \textbf{Params$\downarrow$} \\
		\midrule
		PointPillars~\cite{langPointPillarsFastEncoders2019a} & L-128 & - & PointPillars & 16.6 & 28.69G & 4.86M \\
		PointPillars~\cite{langPointPillarsFastEncoders2019a} & L-32 & - & PointPillars & 33.9 & 27.16G & 4.86M \\
		\midrule
		PointPillars~\cite{langPointPillarsFastEncoders2019a} & R & - & PointPillars & 62.2 & 25.33G & 4.85M \\
		RadarPillarNet~\cite{zhengRCFusionFusing4D2023} & R & - & RadarPillarNet & 60.3 & 25.32G & 4.85M \\
		\midrule
		LSS~\cite{philionLiftSplatShoot2020b} & C & 544$\times$960 & R50 & 3.7 & 2746.61G & 52.91M \\
		BEVFormer-S~\cite{liBEVFormerLearningBirdsEyeView2022} & C & 544$\times$960 & R50 & 11.4 & 363.46G & 42.47M \\
		BEVFormer~\cite{liBEVFormerLearningBirdsEyeView2022} & C & 544$\times$960 & R50 & 11.4 & 363.46G & 42.47M \\
		Sparse4Dv3~\cite{linSparse4DV3Advancing2023} & C & 544$\times$960 & R50 & 15.9 & 447.77G & 47.92M \\
		BEVFormer-S~\cite{liBEVFormerLearningBirdsEyeView2022} & C & 864$\times$1536 & R101-DCN & 4.7 & 891.18G & 63.21M \\
		BEVFormer~\cite{liBEVFormerLearningBirdsEyeView2022} & C & 864$\times$1536 & R101-DCN & 4.7 & 891.18G & 63.21M \\
		\midrule
		BEVFusion~\cite{liuBEVFusionMultiTaskMultiSensor2023a} & R+C & 544$\times$960 & R50+PointPillars & 3.6 & 2858.51G & 60.04M \\
		RCFusion~\cite{zhengRCFusionFusing4D2023} & R+C & 544$\times$960 & R50+RadarPillarNet & 3.6 & 2857.11G & 59.89M \\
		RCBEVDet~\cite{linRCBEVDetRadarCameraFusion2024a} & R+C & 544$\times$960 & R50+RadarBEVNet & 5.2 & - & - \\
		Doracamom-S~\cite{zhengDoracamomJoint3D2026} & R+C & 544$\times$960 & R50+RadarPillarNet & 4.8 & 1842.41G & 55.88M \\
		Doracamom~\cite{zhengDoracamomJoint3D2026} & R+C & 544$\times$960 & R50+RadarPillarNet & 4.2 & 1842.41G & 55.88M \\
		\rowcolor{gray!20} 
		Sparse4D-Radar-Base(\textbf{ours}) & R+C & 544$\times$960 & R50+PointPillars & \textbf{11.5} & \textbf{472.06G} & \textbf{53.47M} \\
		\rowcolor{gray!20} 
		Sparse4D-Radar-Acc(\textbf{ours}) & R+C & 544$\times$960 & R50+PointPillars & \underline{8.7} & \underline{482.96G} & \underline{55.46M} \\		
		\bottomrule
		\multicolumn{7}{l}{\scriptsize \textit{Note:} All models are evaluated on the OmniHD-Scenes test set using a single NVIDIA RTX 4090 GPU.}
	\end{tabular}	
\end{table*}

\textbf{Superiority in Object Detection.} Table~\ref{tab:all_conditions} presents the performance comparison of different methods on the OmniHD-Scenes~\cite{zhengOmniHDScenesNextGenerationMultimodal2026} test set for 3D object detection. Our Sparse4D-Radar framework achieves superior overall performance compared to other methods based on radar-camera fusion. Specifically, our efficiency-oriented Sparse4D-Radar-Base model surpasses the leading method Doracamom~\cite{zhengDoracamomJoint3D2026}, achieving substantial improvements of 7.89\% in mAP and 11.03\% in ODS. The performance gap is even larger when compared to other strong baselines like BEVFusion~\cite{liuBEVFusionMultiTaskMultiSensor2023a} (+13.06\% mAP and +14.25\% ODS), RCFusion~\cite{zhengRCFusionFusing4D2023} (+12.13\% mAP and +15.72\% ODS), RCBEVDet~\cite{linRCBEVDetRadarCameraFusion2024a} (+11.48\% mAP and +11.84\% ODS), and RaGS~\cite{baiRaGSUnleashing3D2025} (+11.13\% mAP and +13.80\% ODS). Furthermore, our accuracy-oriented variant, Sparse4D-Radar-Acc, pushes the performance envelope even further, elevating the detection accuracy to 47.57\% mAP and the overall score to 58.35\% ODS. These consistent improvements across all metrics demonstrate that our sparse fusion architecture, even in its most efficient configuration, offers a superior mechanism for multi-modal feature integration compared to conventional radar–camera fusion methods.

A standout achievement of our framework is the drastic reduction in mAVE across all benchmarked methods. This superiority stems from both physical and architectural advantages: while 4D radar inherently possesses a lower velocity error than camera-only or LiDAR-only methods due to its direct velocity measurements, the efficacy of our specific fusion strategy is most evident when compared to other radar-camera fusion methods. Specifically, Sparse4D-Radar-Base already achieves a remarkably low mAVE of 0.3368, and Sparse4D-Radar-Acc—strengthened by the VCS and AMG modules—establishes a record-breaking mAVE of 0.3187. This exceptional precision suggests that our sparse query-based framework is inherently more capable of preserving and leveraging high-fidelity velocity information, effectively bypassing the feature-blurring effects common in dense BEV-based fusion schemes and fully unleashing the potential of 4D radar for dynamic object perception.

Beyond its superiority within the multi-modal domain, Sparse4D-Radar successfully bridges the performance gap with LiDAR-based systems. Specifically, our framework delivers an ODS that outperforms both the 32-line (52.66\%) and 128-line (55.54\%) LiDAR-based systems. By leveraging the unique physical advantages of 4D radar through an optimized fusion framework, Sparse4D-Radar proves that a cost-effective radar-camera configuration can serve as a high-precision and robust alternative to expensive LiDAR sensors without compromising overall perception quality.

\textbf{Robustness in Adverse Environments.} To further evaluate the robustness of Sparse4D-Radar, we conduct experiments under challenging environmental conditions, including night and rainy weather, as detailed in Table~\ref{tab:bad_conditions}. Adverse weather conditions typically pose a severe threat to vision-dependent systems due to reduced visibility, glare, and motion blur. As evidenced by the results, purely camera-based methods suffer significant performance degradation in these settings. 

In contrast, Sparse4D-Radar maintains remarkably stable performance, effectively leveraging the all-weather robustness of 4D radar to compensate for visual uncertainty. Specifically, our Sparse4D-Radar-Base achieves an mAP of 49.05\% and an ODS of 58.58\%, while the Sparse4D-Radar-Acc further pushes the comprehensive score to a record-breaking 58.61\% ODS. These results represent a substantial lead over the previous multi-modal state-of-the-art approaches. Most notably, our framework achieves the lowest errors across all mTP metrics among all radar-camera methods. This consistent and comprehensive error reduction underscores the architectural efficacy of our sparse fusion approach, which successfully integrates stable geometric priors from radar while maintaining high-fidelity perception even when visual reliability is compromised.

\textbf{Efficiency and Real-time Capability.} Beyond achieving superior detection accuracy, Sparse4D-Radar exhibits significant advantages in computational efficiency and real-time inference capability, as summarized in Table~\ref{tab:computational_cost}. Compared to existing multi-modal fusion methods, Sparse4D-Radar demonstrates superior inference time. Running on a single NVIDIA RTX 4090 GPU, Sparse4D-Radar-Base achieves an inference speed of 11.5 FPS—outperforming both BEVFusion~\cite{liuBEVFusionMultiTaskMultiSensor2023a} (3.6 FPS) and RCFusion~\cite{zhengRCFusionFusing4D2023} (3.6 FPS) by more than 3$\times$, while maintaining a significant lead over Doracamom~\cite{zhengDoracamomJoint3D2026} (4.2 FPS) and RCBEVDet~\cite{linRCBEVDetRadarCameraFusion2024a} (5.2 FPS). Even our accuracy-optimized variant, Sparse4D-Radar-Acc, maintains a stable output of 8.7 FPS, effectively doubling the frame rate of most contemporary methods. The efficiency of our framework is further reflected in its reduced computational overhead. While dense-fusion methods incur a heavy computational burden between 1,842.41G and 2,858.51G FLOPs, our Sparse4D-Radar-Base and Sparse4D-Radar-Acc requires only 472.06G and 482.96G FLOPs, respectively. Furthermore, our models maintain competitive parameter counts (53.47M and 55.46M, respectively), both of which are lower than those of most fusion-based counterparts.

The efficiency of Sparse4D-Radar stems from its sparse query-based architecture. By performing multi-modal sampling and interaction directly on sparse queries, our framework bypasses the memory-intensive LSS~\cite{philionLiftSplatShoot2020b} operation and redundant feature interaction typical of dense-fusion methods. This design effectively alleviates computational bottlenecks, bridging the gap between high-precision perception and real-time on-board deployment.

\subsection{Ablation Study}

\begin{table}[t]
	\centering
	\caption{Ablation study on proposed modules.}
	\label{tab:proposed_modules}
	\begin{tabular}{c|cc|c}
		\toprule
		\textbf{Method} & \textbf{mAP$\uparrow$} & \textbf{ODS$\uparrow$} & \textbf{FPS$\uparrow$} \\
		\midrule
		Baseline (Base) & 47.01 & 57.25 & 11.5 \\
		+VCS & 47.04 (\textcolor{mygreen}{\textbf{0.03}$\uparrow$}) & 57.76 (\textcolor{mygreen}{\textbf{0.51}$\uparrow$}) & 8.9 (\textcolor{myred}{\textbf{2.6}$\downarrow$}) \\
		+AMG & 46.70 (\textcolor{myred}{\textbf{0.31}$\downarrow$}) & 57.31 (\textcolor{mygreen}{\textbf{0.06}$\uparrow$}) & 11.0 (\textcolor{myred}{\textbf{0.5}$\downarrow$}) \\
		+VCS+AMG (Acc) & 47.57 (\textcolor{mygreen}{\textbf{0.56}$\uparrow$}) & 58.35 (\textcolor{mygreen}{\textbf{1.10}$\uparrow$}) & 8.7 (\textcolor{myred}{\textbf{2.8}$\downarrow$}) \\		
		\bottomrule
	\end{tabular}	
\end{table}

\begin{table}[t]
	\centering
	\caption{Ablation study on radar backbone.}
	\label{tab:radar_backbone}	
	\begin{tabular}{c|cc|c}
		\toprule
		\textbf{Method} & \textbf{mAP$\uparrow$} & \textbf{ODS$\uparrow$} & \textbf{FPS$\uparrow$} \\
		\midrule
		PointPillars & \underline{47.01} & \underline{57.25} & \underline{11.5} \\
		Second & \textbf{47.71} & \textbf{57.61} & 9.6 \\
		RadarBEVNet & 46.60 & 56.96 & 1.5 \\
		RadarPillarNet & 44.97 & 56.80 & \textbf{12.1} \\
		\bottomrule
	\end{tabular}
\end{table}

\begin{table}[t]
	\centering
	\setlength{\tabcolsep}{3pt}
	\caption{Ablation study on Feature Sampling.}
	\label{tab:feature_sampling}	
	\begin{tabular}{c|cc|cccc}
		\toprule
		\textbf{Method} & \textbf{mAP$\uparrow$} & \textbf{ODS$\uparrow$} 
		& \textbf{mATE$\downarrow$} & \textbf{mASE$\downarrow$} & \textbf{mAOE$\downarrow$} & \textbf{mAVE$\downarrow$} \\
		\midrule
		w/o Offset & \underline{47.01} & \underline{57.25} & 0.4388 & 0.1981 & 0.3266 & 0.3368 \\
		Shared Offset & 46.61 & 57.18 & 0.4434 & \underline{0.1961} & 0.3125 & 0.3380 \\
		Level-wise Offset & 44.08 & 56.22 & \textbf{0.4280} & \textbf{0.1945} & \underline{0.3106} & \underline{0.3330} \\
		VCS  & \textbf{47.04} & \textbf{57.76} & \underline{0.4333} & \underline{0.1961} & \textbf{0.3081} & \textbf{0.3234} \\
		\bottomrule
	\end{tabular}	
\end{table}

\begin{table}[t]
	\centering
	\setlength{\tabcolsep}{3pt}
	\caption{Ablation study on Multimodal Fusion.}
	\label{tab:multimodal_fusion}	
	\begin{tabular}{c|cc|cccc}
		\toprule
		\textbf{Method} & \textbf{mAP$\uparrow$} & \textbf{ODS$\uparrow$} 
		& \textbf{mATE$\downarrow$} & \textbf{mASE$\downarrow$} & \textbf{mAOE$\downarrow$} & \textbf{mAVE$\downarrow$} \\
		\midrule
		Concat & \textbf{47.01} & \underline{57.25} & \underline{0.4388} & \underline{0.1981} & 0.3266 & \underline{0.3368} \\
		Attention & 40.22 & 53.43 & 0.4544 & 0.2002 & \underline{0.3261} & 0.3540 \\
		AMG & \underline{46.70} & \textbf{57.31} & \textbf{0.4367} & \textbf{0.1970} & \textbf{0.3164} & \textbf{0.3332} \\
		\bottomrule
	\end{tabular}	
\end{table}

\textbf{Ablation Study on Proposed Modules.} To verify the effectiveness of the key components in our newly proposed framework, we conduct a comprehensive ablation study. Given the architectural novelty, we establish a robust Baseline using our Sparse4D-Radar-Base configuration, which already incorporates the proposed Deformable Fusion mechanism. This ensures that any observed performance gains are strictly attributable to the newly introduced modules. The results are summarized in Table~\ref{tab:proposed_modules}.

The integration of the VCS module individually brings a notable improvement in detection quality, with ODS increasing from 57.25\% to 57.76\% (+0.51\%). Although the mAP shows a small gain (+0.03\%), the significant boost in ODS suggests that VCS effectively leverages radar's velocity information to refine the spatial alignment of 3D objects. The computational overhead of VCS leads to a drop in inference speed to 8.9 FPS, which is a justifiable trade-off for the enhanced precision in dynamic scenarios.

When applying AMG alone, we observe a slight performance fluctuation. ODS increases by 0.06\%, while mAP decreases by 0.31\%. This indicates that without the precise features, the attribute-based guidance might introduce some noise or over-constrain the feature learning process. However, AMG maintains a high execution efficiency, only reducing the speed by 0.5 FPS.

The most compelling results are achieved when both modules are combined, denoted as Sparse4D-Radar-Acc. This configuration yields the best performance across all accuracy metrics, achieving 47.57\% mAP (+0.56\%) and 58.35\% ODS (+1.10\%). Interestingly, the synergistic effect of the two modules far exceeds their individual contributions. The precision gained from VCS's velocity filtering allows AMG to operate on more reliable feature representations, leading to a substantial leap in overall detection robustness. Although the final FPS is 8.7, the significant gains in mAP and ODS demonstrate the superiority of our complete framework for high-precision autonomous driving tasks.

\textbf{Ablation Study on Radar Backbone.} To strike a balance between detection accuracy and inference efficiency, we evaluate several radar backbones, as shown in Table~\ref{tab:radar_backbone}. While Second~\cite{yanSECONDSparselyEmbedded2018a} achieves the highest precision (47.71\% mAP), its lower speed (9.6 FPS) poses a challenge for real-time applications. Conversely, RadarPillarNet~\cite{zhengRCFusionFusing4D2023} offers the highest efficiency (12.1 FPS) but suffers from a significant drop in mAP (44.97\%).

We ultimately select PointPillars as our default backbone because it offers the most favorable trade-off: it maintains a competitive detection precision (47.01\% mAP and 57.25\% ODS) that is closely comparable to Second, while significantly enhancing inference speed to 11.5 FPS. This choice ensures that our framework preserves high-quality perception capabilities while maximizing real-time processing throughput.

\textbf{Ablation Study on Feature Sampling.} To mitigate spatial misalignment caused by imperfect temporal synchronization, calibration errors, or inherent sensor noise, we investigate several strategies during the feature sampling stage. The quantitative results are presented in Table~\ref{tab:feature_sampling}. Compared with the version without offset, we first try two learnable offset methods. Shared Offset introduces shared offsets across different feature levels; however, this level-insensitive approach fails to capture the multi-level geometric nuances, leading to a slight performance drop. In contrast, Level-wise Offset employs level-sensitive offsets to refine sampling locations. While it achieves the lowest mATE and mASE, the intensive memory requirements and high computational overhead significantly hinder its practical deployment.

Our proposed VCS achieves the best overall performance, reaching 47.04\% mAP and 57.76\% ODS. Notably, VCS significantly outperforms previous methods in motion-related metrics, yielding the lowest errors in orientation (0.3081 mAOE) and velocity (0.3234 mAVE). This success stems from its good use of 4D radar's velocity , which effectively rectifies spatial misalignments across sensors. As a low-parameter approach, VCS provides a robust and computationally efficient solution that minimizes memory overhead, demonstrating exceptional effectiveness and hardware-friendliness for real-time multi-modal feature alignment.

\begin{figure*}[t]
	\centering
	\includegraphics[width=1.0\linewidth]{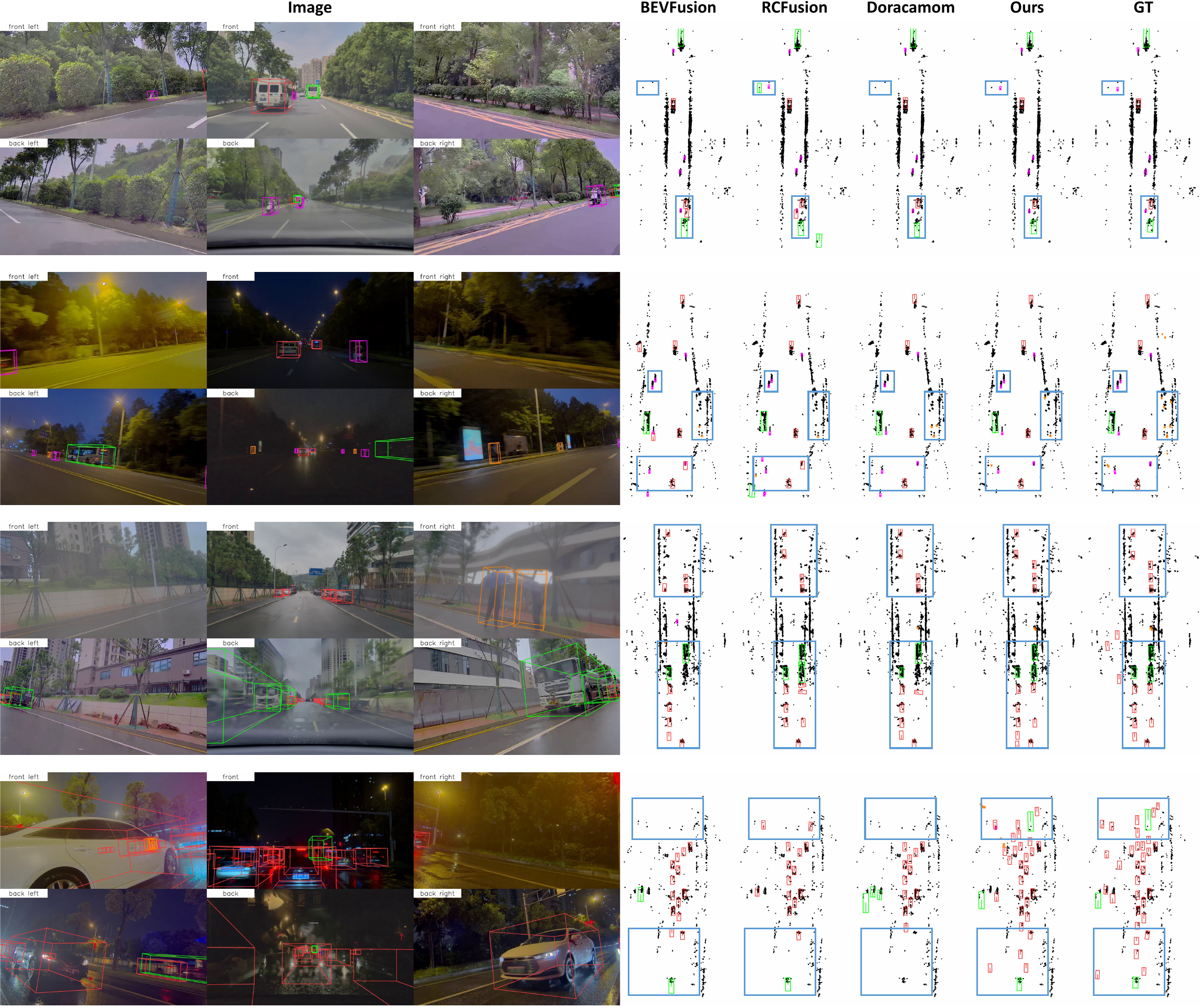}	
	\caption{Visualization results for 3D object detection on the OmniHD-Scenes test set. From top to bottom, each row corresponds to clear daytime, clear nighttime, rainy daytime, and rainy nighttime scenarios, respectively. The left column shows our predicted 3D bounding boxes projected onto multi-view images. Different object categories (\textcolor{car}{car}, \textcolor{pedestrian}{pedestrian}, \textcolor{rider}{rider}, and \textcolor{largevehicle}{large\_vehicle}) are distinguished by color. The right column presents a comparison among BEVFusion, RCFusion, Doracamom, our method, and the ground truth. In the BEV view, 4D radar points are represented in black, and highlighted regions indicate where our method more effectively reduces false positives and missed detections compared to other approaches.}
	\label{fig:visualization}
\end{figure*}

\textbf{Ablation Study on Multimodal Fusion.} To identify the most effective way for fusing multi-modal features, as shown in Table~\ref{tab:multimodal_fusion}, we compare three representative fusion strategies. We first consider Concat, which serves as a straightforward baseline by directly stacking multi-modal features along the channel dimension. This method yields the highest mAP of 47.01\%, demonstrating a robust capacity to preserve the original informational density from both modalities. Alternatively, we evaluate a more complex strategy named Attention~\cite{vaswaniAttentionAllYou2017b}, which utilizes cross-attention to perform mutual querying between multi-modal features. Contrary to expectations, despite its theoretical potential for high-capacity modeling, it exhibits the lowest performance (40.22\% mAP) and entails heavy computational costs. The quadratic complexity of global querying fails to translate into superior performance, demonstrating that high-capacity modeling does not inherently guarantee effective multi-modal fusion in this context.

In contrast, our proposed AMG module achieves a superior balance between overall metrics and attribute error. Although its mAP (46.70\%) is marginally lower than Concat, it outperforms all other methods across ODS (57.31\%), mATE (0.4367), mASE (0.1970), mAOE (0.3164), and mAVE (0.3332). These improvements highlight the effectiveness of AMG in adaptively modulating multi-modal contributions. By selectively suppressing noise-prone features, AMG promotes a more robust and reliable fusion representation, leading to notable gains in 3D object detection performance.

\subsection{Visualization Results}

To qualitatively evaluate the effectiveness of the proposed method, we present visualization results on the OmniHD-Scenes~\cite{zhengOmniHDScenesNextGenerationMultimodal2026} test set under diverse environmental conditions. As shown in Fig.~\ref{fig:visualization}, our approach demonstrates robust perception performance across varying illumination and weather scenarios, including clear daytime, clear nighttime, rainy daytime, and rainy nighttime. Compared with BEVFusion~\cite{liuBEVFusionMultiTaskMultiSensor2023a}, RCFusion~\cite{zhengRCFusionFusing4D2023}, and Doracamom~\cite{zhengDoracamomJoint3D2026}, our method produces more accurate and consistent 3D bounding box predictions, particularly in challenging cases with sparse observations or severe noise interference.

In the multi-view image projections, our predictions exhibit improved localization accuracy and tighter alignment with object boundaries. From the BEV perspective, the integration of 4D radar information enables our model to better capture spatial structure and motion cues, leading to enhanced detection reliability. Notably, in the highlighted regions, competing methods tend to suffer from false positives or missed detections, whereas our approach effectively suppresses spurious responses while preserving true targets.

These results indicate that the proposed fusion framework can more effectively leverage complementary information from different modalities, resulting in improved robustness and generalization in complex real-world driving scenarios.

\section{Conclusion}
\label{sec:con}

In this paper, we propose Sparse4D-Radar, a novel 3D object detection framework designed to bridge the gap in surround-view radar-camera fusion for autonomous driving. At its core, a Deformable Fusion module seamlessly integrates 4D radar and camera features using a sparse-query mechanism. To further improve localization precision and modality reliability, the framework incorporates two specialized components: the VCS module, which leverages radar velocity measurements for point feature refinement, and the AMG module, which dynamically regulates multi-modal integration based on feature reliability. Extensive experiments demonstrate that Sparse4D-Radar achieves state-of-the-art performance, delivering a superior balance between detection accuracy and computational efficiency.

Despite these advances, certain limitations remain. While additional modules can enhance accuracy, they introduce computational overhead that complicates real-time deployment. Furthermore, the framework does not fully exploit the inherent characteristics of 4D radar, and radar feature extraction remains suboptimal. Future work will address these issues.

\bibliographystyle{IEEEtran}
\bibliography{ref.bib}

\newpage

\begin{IEEEbiography}[{\includegraphics[width=1in,height=1.25in,clip,keepaspectratio]{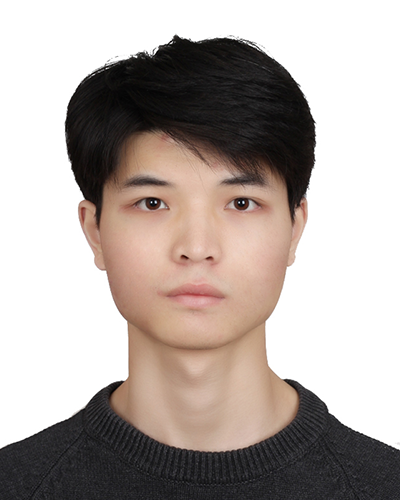}}]{Fuyuan Ai} was born in Yichang, Hubei, China, in 1997. He received the B.S. degree in Ocean College from Zhejiang University, Hangzhou, China, in 2020. He is pursuing the Ph.D. degree in marine technology and engineering from Zhejiang University, Hangzhou, China.
	
	His research interests are multi-sensor fusion perception based on 4D radar, especially point cloud generation and object detection.
\end{IEEEbiography}

\begin{IEEEbiography}[{\includegraphics[width=1in,height=1.25in,clip,keepaspectratio]{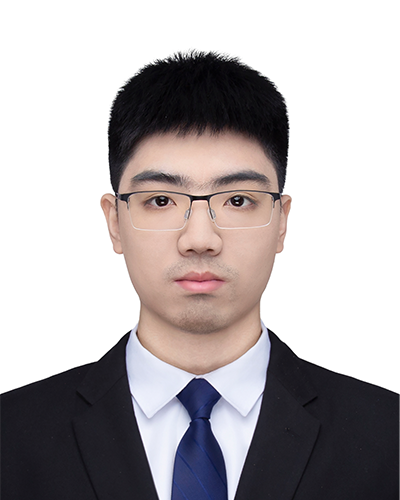}}]{Yuchen Tan} received the B.S. degree in Ocean Engineering and Technology from Zhejiang University, Hangzhou, China, in 2023. He is currently pursuing the Ph.D. degree in Marine Technology and Engineering at Zhejiang University, Hangzhou, China.
	
	His research interests include intelligent perception and sensor fusion.
\end{IEEEbiography}

\begin{IEEEbiography}[{\includegraphics[width=1in,height=1.25in,clip,keepaspectratio]{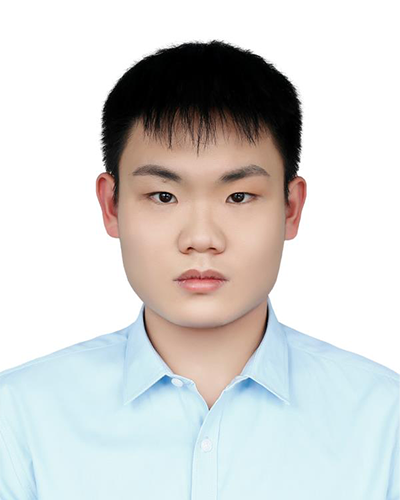}}]{Jiehui Chen} received the B.S. degree in automation from Northeastern University, Shenyang, China, in 2022. He is currently pursuing the M.S. degree in electronic information with the Ocean College, Zhejiang University, Zhoushan, China.
	
	His current research interests include 4D millimeter-wave radar, image processing, fusion perception, and radar signal processing.
\end{IEEEbiography}

\begin{IEEEbiography}[{\includegraphics[width=1in,height=1.25in,clip,keepaspectratio]{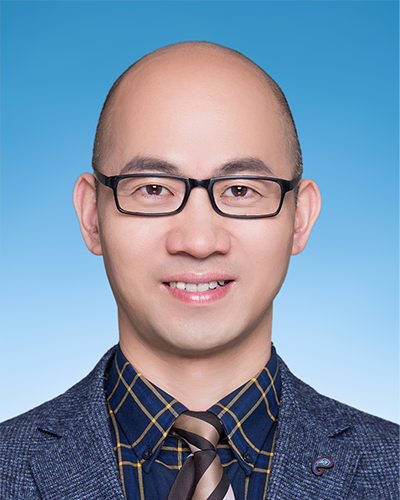}}]{Chunyi Song} (Member, IEEE) received the Ph.D. degree in electronic and communication engineering from Waseda University, Tokyo, Japan, in 2007.
	
	From 2007 to 2009, he was a Research Associate with Waseda University. He was with the National Institute of Information and Communications Technology (NICT), Koganei, Japan, as a Researcher, from 2009 to 2013, and as a Senior Researcher, in 2014. Since 2014, he has been with Zhejiang University, Hangzhou, China, as an associate professor, where now he is a full professor and also the Executive Vice Director of the Engineering Research Center of Oceanic Sensing Technology and Equipment, Ministry of Education. Since 2022, he has also been with the Donghai Laboratory, Zhoushan, China, as a principle scientist.
\end{IEEEbiography}

\begin{IEEEbiography}[{\includegraphics[width=1in,height=1.25in,clip,keepaspectratio]{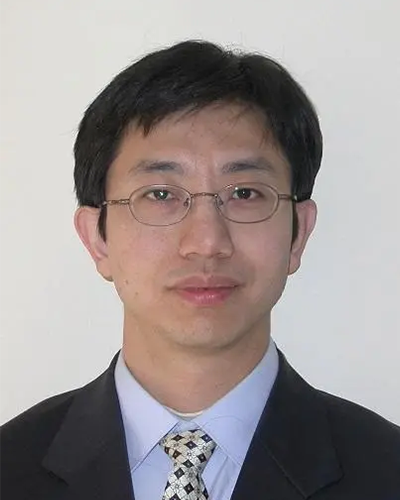}}]{Zhiwei Xu} (Senior Member, IEEE) received the Ph.D. degree in integrated circuits and systems from the University of California at Los Angeles, Los Angeles, CA, USA, in 2003.
	
	He held senior manager positions at G-Plus Inc., SST Communications, Conexant Systems, and NXP Inc., before he joined HRL Laboratories, where he led development for wireless LAN and SoC solutions for proprietary wireless multimedia systems, CMOS cellular transceiver, multimedia over cable (MoCA) systems, and TV tuners. He is a Professor with Zhejiang University, Hangzhou, China, working on cognitive radios, high-speed ADC, and millimeter-wave (MMW) integrated circuits (ICs). He has authored or co-authored 70 publications, one contribution to the encyclopedia of wireless and mobile communications, ten granted patents, and around 20 pending ones.
	
	Dr. Xu was elected to the Distinguished Expert, China, in 2014, and the Team Leader of the Leading Innovative Team of Zhejiang Province, in 2018. He serves on the Technical Program Committee for International Workshop on MMW Communications: From Circuit to Network. He serves as a Series Editor for the IEEE Communications Magazine.
\end{IEEEbiography}

\end{document}